\documentclass[fleqn,10pt]{wlscirep}
\usepackage[utf8]{inputenc}
\usepackage[T1]{fontenc}
\usepackage{color}
\usepackage{booktabs}
\usepackage{wrapfig}
\usepackage{caption}
\title{Perfectly predicting ICU length of stay: too good to be true}

\author[1,*]{Sandeep Ramachandra}
\author[1]{Gilles Vandewiele}
\author[1]{David Vander Mijnsbrugge}
\author[1]{Femke Ongenae}
\author[1]{Sofie Van Hoecke}

\affil{IDLab, Ghent University -- imec, Ghent, Belgium}

\affil[*]{sandeep.ramachandra@ugent.be}

\begin{document}
\flushbottom
\maketitle
\thispagestyle{empty}

A paper of Alsinglawi et al.~\cite{Alsinglawi2022Jan} was recently accepted and published in Scientific Reports. In this paper, the authors aim to predict length of stay (LOS), discretized into either long ($> 7$ days) or short stays ($\leq 7$ days), of lung cancer patients in an ICU department using various machine learning techniques. The authors claim to achieve perfect results with an Area Under the Receiver Operating Characteristic curve (AUROC) of $100\%$ with a Random Forest (RF) classifier with ADASYN class balancing over sampling technique, which if accurate could have significant implications for hospital management. However, we have identified several methodological flaws within the manuscript which cause the results to be overly optimistic and would have serious consequences if used in a clinical practice. Moreover, the reporting of the methodology is unclear and many important details are missing from the manuscript, which makes reproduction extremely difficult. We highlight the effect these oversights have had on the result and provide a more believable result of 88.91\% AUROC when these oversights are corrected. \\

The dataset used within the study is small with 119 lung cancer patients extracted from the \textsc{mimic-iii} dataset. This raises concerns regarding \textbf{significance} of any of the conclusions made within this manuscript. A decision was made to discretize the LOS into two bins (15 long stays and 104 short stays), rather than regression prediction or survival analysis techniques~\cite{ohno2001modeling} which would be more informative for ICU capacity management. This results in a \textbf{loss of information and a decrease of clinical usefulness}. In the supplementary information, S8.2, RF classifier are reported to perform best on the dataset without any under- or oversampling. However, an accuracy of $87.4\%$ is reported for this best-performing model, which corresponds to the fraction of the majority class ($\frac{104}{104+15}$). As such, it seems that the \textbf{model has not actually learned much more than to always predict the majority class}. \\

\begin{wrapfigure}{l}{0.42\textwidth}
    \centering
    \includegraphics[width=0.55\linewidth]{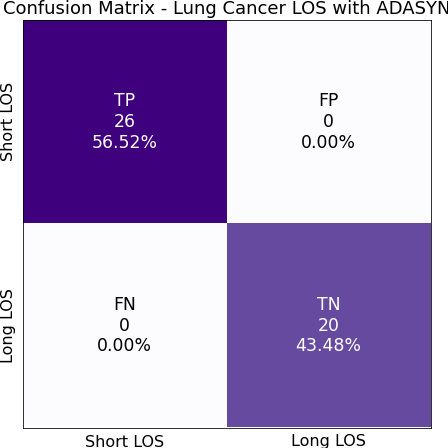}
    \caption{Figure 3 from Alsinglawi et al.~\cite{Alsinglawi2022Jan}. The confusion matrix shows that the long LOS contains synthetic samples as the number of samples exceed those present in the original dataset(15).\href{https://creativecommons.org/licenses/by/4.0/}{CC BY 4.0}}
    \label{fig:confusion_matrix}
\end{wrapfigure}

The data retrieved from the \textsc{mimic-iii} database is extremely imbalanced as only $12.6\%$ ($\frac{15}{119}$) of the patients stayed longer than seven days. To combat this data imbalance, the authors propose to use oversampling techniques. Upon looking at the \href{https://github.com/balsinglawi/predict-lung-cancer-LOS}{code provided by the authors}, it appears as though the authors have split performed a 70\%/30\% split before class balancing and then split the class balanced by 70\%/30\% to get their train and test set. Since \textbf{both} the train and test sets have been over sampled, the models have \textbf{highly optimistic results} that merely reflect the their capability to memorize \textbf{test samples seen during training}, rather than its predictive performance~\cite{Vandewiele2021Jan}. This is clear from Figure \ref{fig:confusion_matrix}, where confusion matrices are presented that correspond to the reported predictive performance scores. However, these confusion matrices clearly contain synthetic samples generated through oversampling, as there are more long LOS samples (20) than in the actual dataset (15). \\

The authors have made use of measurements, demographics information, and medication information. Much of this is only available after the patient spends some time in the ICU. This has negative implications on the clinical usefulness, as the data might not be available at admission time. Moreover, severe selection bias is introduced by removing any patients who died \emph{during} their admission. \\

We set out to reproduce our suspicion of imputation and oversampling  before data partitioning being the main contributor for the highly optimistic results. The authors have unfortunately not made the code publicly available to retrieve these features, and were not willing to share the code upon our request. To overcome this, we tried to mimic the data used within the paper~\cite{Alsinglawi2022Jan} as closely as possible. While authors have given supplementary information with instructions on reproducing their dataset, the instructions lack important details. For example, instructions to exclude all non cancer admissions distinctly by each patient are provided, but it is not specified how these exclusion criteria are achieved. Other challenges include: (i) absence of origin for each feature, i.e. the \textsc{csv} file they were drawn from, (ii) features not matching the \textsc{mimic-iii} drug name / lab name, and (iii) some features not being present in the \textsc{mimic-iii} data. Due to these challenges, our queries can only mimic as best as possible the original dataset. The steps to get the dataset are: 
\begin{itemize}
    \item From the \texttt{ADMISSIONS} file, remove all \emph{admissions} with \texttt{EXPIREFLAG} equal to 1.
    \item From \texttt{ADMISSIONS} file, exclude all \emph{subjects} whose \texttt{DIAGNOSIS} does not contain the word ``cancer'', and also those missing a \texttt{HADM\_ID} or \texttt{ICUSTAY\_ID} (taken from \texttt{ICUSTAYS} file).
    \item For the remaining \texttt{SUBJECT\_ID}, retain patients diagnosed with lung cancer, using the \textsc{icd-9} codes (162.x).
    \item Using these \texttt{SUBJECT\_ID}, we can fetch the medications (see list in Supplementary document S2.1\cite{Alsinglawi2022Jan}) from \texttt{DRUG} column in the \texttt{PRESCRIPTIONS} data using a list of \texttt{ICUSTAY\_ID} for each patient as a binary feature (1 if they were administered the medication, 0 otherwise). The general patient information , such as the LOS, gender, age category (1 if age $>$ 60 for last admission 0 otherwise) and admission type, can be gathered from \texttt{ADMISSIONS} and \texttt{ICUSTAYS} files for the latest \texttt{HADM\_ID} and \texttt{ICUSTAY\_ID}. The remaining features are all labs taken from \texttt{CHARTEVENTS} file. These are reduced to the mean for each feature for each patient.
\end{itemize}
\begin{table}
    \centering
    \begin{tabular}{ccc}
    \toprule
         & \begin{tabular}{@{}c@{}}Reported\\dataset\end{tabular} &\begin{tabular}{@{}c@{}}Our\\dataset\end{tabular} \\ \midrule
        Total patients & 119 & 112 \\
        \begin{tabular}{@{}c@{}}long LOS ($> 7$ days)\end{tabular}  & 15 & 10 \\ \bottomrule 
    \end{tabular}
    \captionsetup{justification=centering}
    \caption{Number of patient samples obtained from \textsc{mimic-iii} dataset}
    \label{tab:Patient_count}
\end{table}

Moreover, it should be noted that the medications and lab keys are all provided without spaces and in lowercase. However, some keys are present in both medications and in labs and they may have suffixes, which raises further ambiguity. As seen in Table \ref{tab:Patient_count}, our data serves as an approximation of the data used in the paper due to the aforementioned difficulties, though we are aware that more long stay patients are missing which could affect our results.\\

Once we retrieved the data, we partitioned it using stratified cross-validation with 10 splits (not performed by the authors), which equals the number of positive samples. Three setups were applied: (i) imputation and oversampling after partitioning into train and test data, (ii) no oversampling but imputation based on train data only, and (iii) oversampling and imputation on the entire dataset, before splitting it into train and test. For each setup, we report the mean AUROC over the folds and its corresponding standard deviation. \\

\begin{table}[h!]
\centering
\begin{tabular}{lc} \toprule
Method & AUROC (in \%) \\ \midrule
(i) imputation + oversampling \emph{after} partitioning & $88.91 \pm 24.97$ \\
(ii) \emph{no} oversampling & $86.73 \pm 16.57$ \\
(iii) imputation + oversampling \emph{before} partitioning & $99.90 \pm 00.30$ \\ \bottomrule
\end{tabular}
\caption{Reproduced results using different setups.}
\label{tab:Model_results}
\end{table}

As seen from Table \ref{tab:Model_results}, we are able to closely replicate the perfect 100\% AUROC reported by the authors when doing the oversampling wrong, leading to overly optimistic results. We also show that when the imputation and oversampling is performed after the partitioning, as it should have been correctly done, the results drop down by 11\%. Additionally, we felt the author's paper would have benefited from an ablation study showing the advantage of oversampling.

\section*{Data Availability} \label{data}
The \textsc{mimic-iii} dataset is available using instructions at \url{https://physionet.org/content/mimiciii/1.4/}. All the code used in the paper along with the instructions on using it can be found at \url{https://github.com/predict-idlab/class-balancing-paper}
\bibliography{sample.bib}

\section*{Author Contributions}
Sandeep Ramachandra and Gilles Vandewiele planned and executed the code for the reproduction of the paper and wrote the manuscript for the commentary. David Vander Mijnsbrugge, Femke Ongenae, and Sofie Van Hoecke reviewed and approved the final draft of the commentary.

\section*{Competing Interests}
The authors declare no competing interests.

\end{document}